\pdfoutput=1

\documentclass[11pt]{article}

\usepackage[]{acl}

\usepackage{times}
\usepackage{latexsym}
\usepackage[super]{nth}

\usepackage[T1]{fontenc}

\usepackage[utf8]{inputenc}

\usepackage{microtype}

\title{The Arabic Parallel Gender Corpus 2.0:\\ Extensions and Analyses}

\author{Bashar Alhafni, Nizar Habash, Houda Bouamor\textsuperscript{\textdagger} \\
  Computational Approaches to Modeling Language Lab\\
  New York University Abu Dhabi \\
  \textsuperscript{\textdagger}Carnegie Mellon University in Qatar\\
  \texttt{\{alhafni,nizar.habash\}@nyu.edu},
  \texttt{hbouamor@qatar.cmu.edu}
  }

\usepackage{graphicx}
\usepackage{url}
\usepackage{latexsym}
\usepackage{makecell}
\usepackage{epstopdf}
\usepackage{tipa}
\usepackage{hyperref}
\usepackage{xstring}
\usepackage{amsfonts}
\usepackage{amsmath}
\usepackage{fixltx2e}
\usepackage[T1]{fontenc}
\usepackage{booktabs}
\usepackage{arabtex}
\usepackage{hhline}
\usepackage{pdfpages}
\usepackage{utf8}
\newcommand{\hide}[1]{}

\newcommand{\AMADDA}{{\={A}}}
\newcommand{\AHAMZAUP}{{\^{A}}}

\newcommand{\AHAMZADN}{{\v{A}}}
\newcommand{\YHAMZA}{{\^{y}}}
\newcommand{\TAMARBUTA}{{$\hbar$}}

\newcommand{\AYN}{{$\varsigma$}}

\newcommand{\AMAQSURA}{{\'{y}}}

\usepackage{multirow}
\usepackage{subfigure}
\usepackage{caption}
\setcode{utf8}
\vocalize

\begin{document}
\maketitle
\begin{abstract}
Gender bias in natural language processing (NLP) applications, particularly machine translation, has been receiving increasing attention.
Much of the research on this issue has focused on mitigating gender bias in English NLP models and systems. Addressing the problem in poorly resourced, and/or morphologically rich languages has lagged behind, largely due to the lack of datasets and resources. 
In this paper, we introduce a new corpus for gender identification and rewriting in contexts involving one or two target users (I and/or You) -- first and second grammatical persons with independent grammatical gender preferences. We focus on Arabic, a gender-marking morphologically rich language. The corpus has  multiple parallel components: four combinations of \nth{1} and \nth{2} person in feminine and masculine grammatical genders, as well as English, and English to Arabic machine translation output.
This corpus expands on \newcite{habash-etal-2019-automatic}'s Arabic Parallel Gender Corpus (APGC~v1.0) by adding second person targets as well as increasing the total number of sentences over 6.5 times, reaching over 590K words.
Our new dataset will aid the research and development of gender identification, controlled text generation, and post-editing rewrite systems that could be used to personalize NLP applications and provide users with the correct outputs based on their grammatical gender preferences. We make the Arabic Parallel Gender Corpus (APGC~v2.0)  publicly available.\footnote{The corpus is available through the CAMeL Lab Resources page: \url{http://resources.camel-lab.com/}}
\end{list} 
\end{abstract}


\section{Introduction}
The great recent advances in many NLP applications have raised expectations about their end users' experiences, particularly in regards to gender identities. Gender negative and positive stereotypes are manifest in most of the world's languages~\cite{maass1996language,menegatti2017gender} and are propagated and amplified by NLP systems~\cite{sun-etal-2019-mitigating}, which not only degrades users' experiences but also creates representational harms~\cite{blodgett-etal-2020-language}. Although human-generated data used to build these systems is considered the main source of these biases, balancing and debiasing the training data do not always lead to less biased systems~\cite{habash-etal-2019-automatic}. This is because the majority of NLP systems are designed to generate a single text output without considering any target user gender information. Therefore, to prevent this and to provide the correct user-aware output, NLP systems should incorporate their users' grammatical gender preferences when available.
Of course, this becomes more challenging for systems targeting multi-user contexts (first, second, and third persons, with independent grammatical gender preferences). One example of this phenomenon is the machine translation of the sentence \textit{I am a doctor and you are a nurse}. While English uses gender neutral terms leading to ambiguous gender references for the first and second persons (\textit{I/doctor} and \textit{you/nurse}), some morphologically rich languages use gender-specific terms for these two expressions. In Arabic, gender-unaware single-output machine translation from English often results in \<أنا طبيب وأنت ممرضة> \textit{{\AHAMZAUP}nA Tbyb w{\AHAMZAUP}nt mmrD{\TAMARBUTA}}\footnote{Arabic transliteration is in the HSB scheme \cite{Habash:2007:arabic-transliteration}.} `I~am~a [male] doctor and you are a [female] nurse', which is inappropriate for female doctors and male nurses, respectively. Alternatively, gender-aware personalized NLP systems should be designed to produce outputs that are as gender-specific as the user information they have access to. Users information could be either embedded as part of the input (e.g., `she is a doctor and he is a nurse') or provided externally by the users themselves.

Aside of context complexity, there is a lack of resources and datasets for morphologically rich languages, where multi-user expressed differences are ubiquitous. In this paper, we focus on Arabic, a gender-marking morphologically rich language. We introduce a new parallel corpus for gender identification and rewriting in contexts involving one or two users -- first and second grammatical persons with independent grammatical gender preferences -- I only, you only, \textit{and} I and You. 
This corpus expands on \newcite{habash-etal-2019-automatic}'s Arabic Parallel Gender Corpus (APGC~v1.0) by adding second person targets as well as increasing the total number of sentences over 6.5 times, reaching over 590K words.
The Arabic Parallel Gender Corpus (APGC~v2.0) also  has  multiple parallel components: four combinations of \nth{1} and \nth{2} person in feminine and masculine grammatical genders, as well as English, and English to Arabic machine translation output.

We make this data publicly available in the hope that it will encourage research and development of gender identification, controlled generation, and post-editing rewrite systems that could be used to personalize NLP applications and provide users with the correct outputs based on their grammatical gender preferences. 
While the work focuses on Arabic, we believe many insights and ideas are easily extensible to other languages, given their linguistics requirements.


This paper is structured as follows. We first discuss some related work (\S\ref{sec:related-work}) and then give a background on Arabic linguistics facts (\S\ref{sec:arabic-facts}). We describe the selection process and the annotation guidelines of our newly created corpus in \S\ref{sec:corpus}. We then present an overview and analysis of our corpus in \S\ref{sec:overview-and-analysis}. Lastly, we show how our corpus could be used to study gender bias in commercial machine translation systems (\S\ref{sec:machine-translation}) and we conclude in \S\ref{sec:conclusion}.  

\section{Related Work}
\label{sec:related-work}
Several approaches have been proposed to mitigate gender bias in various NLP tasks including machine translation \cite{rabinovich-etal-2017-personalized,elaraby2018,vanmassenhove-etal-2018-getting,escude-font-costa-jussa-2019-equalizing,stanovsky-etal-2019-evaluating,costa-jussa-de-jorge-2020-fine,gonen-webster-2020-automatically,saunders-byrne-2020-reducing, saunders-etal-2020-neural,stafanovics2020mitigating,saunders2021worst,savoldi2021gender,ciora-etal-2021-examining}, dialogue systems \cite{cercas-curry-etal-2020-conversational,dinan-etal-2020-queens,liu-etal-2020-gender,liu-etal-2020-mitigating,sheng2021nice,sheng2021revealing}, language modeling \cite{lu2018gender,bordia-bowman-2019-identifying,sheng-etal-2019-woman,vig2020}, co-reference resolution \cite{rudinger-etal-2018-gender,zhao-etal-2018-gender}, and named entity recognition \cite{mehrabi2019man}. The majority of these approaches focus either on debiasing word embeddings (contextualized or non-contextualized) before using them in downstream tasks \cite{bolukbasi2016man,zhao-etal-2018-learning,gonen2019lipstick,manzini2019black,zhao2020gender}, adding additional information to the input to enable models to capture gender information correctly \cite{vanmassenhove-etal-2018-getting,moryossef-etal-2019-filling,stafanovics2020mitigating,saunders-etal-2020-neural}, or creating gender-balanced corpora through counterfactual data augmentation techniques \cite{lu2018gender,hall-maudslay-etal-2019-name,zmigrod-etal-2019-counterfactual}. In terms of rewriting, \newcite{vanmassenhove2021neutral} and \newcite{sun2021they} recently presented rule-based and neural rewriting models to generate gender-neutral sentences.

When it comes to morphologically rich languages, \newcite{vanmassenhove-monti-2021-gender} introduced an English-Italian dataset where the English sentences are gender annotated at the word-level and paired with multiple gender alternative Italian translations when needed. For Arabic, \newcite{habash-etal-2019-automatic} created APGC~v1.0
-- a parallel corpus of first-person-singular Arabic sentences that are gender-annotated and reinflected. They selected the sentences from a subset of the English-Arabic OpenSubtitles 2018 dataset \cite{lison-tiedemann-2016-opensubtitles2016}. Each sentence is labeled based on the grammatical gender of its singular speaker as F (feminine), M (masculine), or B (ambiguous). For the M and F sentences, they introduced their parallel opposite gender forms. Moreover, they developed a two-step model to do gender-identification and reinflection. 
They demonstrated the effectiveness of their approach by applying it to the output of a gender-unaware machine translation system to produce gender-specific outputs. In the same line of work, \newcite{alhafni-etal-2020-gender} used APGC~v1.0 to create a joint gender identification and reinflection sequence-to-sequence model. They treated the problem as a user-aware grammatical error correction task and showed improvements over \newcite{habash-etal-2019-automatic}'s system.

Our work expands
APGC~v1.0 by including contexts involving first and second grammatical persons covering singular, dual, and plural constructions; and we add six times more sentences.

\section{Arabic Linguistic Background}
\label{sec:arabic-facts}
We provide background on the two main challenges that face Modern Standard Arabic (MSA) NLP systems when it comes to gender expressions:  morphological richness and orthographic ambiguity.
\paragraph{Morphological Richness} Arabic is a morphologically rich language that inflects for gender, number, person, case, state, aspect, mood and voice, in addition to various attachable clitics such as prepositions, particles, and pronouns \cite{Habash:2010:introduction}. Gender in Arabic has two values: masculine (\textit{M}) or feminine (\textit{F}), whereas number has three values: singular (\textit{S}), dual (\textit{D}), and plural (\textit{P}). Gender and number apply to verbs, nouns, and adjectives. They are commonly expressed using inflectional suffixes that represent some number and gender combination (for  nominative indefinite): \<ة>+~+{\TAMARBUTA}~(\textit{FS}), \<ان>~+\textit{An}~(\textit{MD}), \<تان>+~+\textit{tAn}~(\textit{FD}), 
\<ون>+~+\textit{wn}~(\textit{MP}), and 
\<ات>+~+\textit{At}~(\textit{FP}). 
For instance, the noun 
\<ممرض>~\textit{mmrD}~`nurse'~(\textit{MS}) could have the following forms:
\<ممرضة>~\textit{mmrD{\TAMARBUTA}}~(\textit{FS}),
\<ممرضان> \textit{mmrDAn}~(\textit{MD}), 
\<ممرضتان>~\textit{mmrDtAn}~(\textit{FD}), 
\<ممرضون> \textit{mmrDwn}~(\textit{MP}), and 
\<ممرضات>~\textit{mmrDAt}~(\textit{FP}). 
Additionally, Arabic has many idiosyncratic templatic stem changes and inflectional suffixes that are not consistent in indicating a specific gender and number combination \cite{alkuhlani-habash-2011-corpus}. In such cases, the functional (grammatical) gender and number do not much the form-based (morphemic) gender and number. 
One example of the so-called {\it Broken Plurals} in Arabic demonstrate this well: 
the plural of  \<عبقري> \textit{{\AYN}bqry} `genius [m.sg]',
\<عباقرة> \textit{{\AYN}bAqr{\TAMARBUTA}} `geniuses [m.pl]',
has a feminine singular suffix but is a masculine plural noun. 
Similarly, the word \<خليفة> \textit{xlyf\TAMARBUTA} `caliph'  is masculine but uses a feminine suffix. 

Gender and number in Arabic participate in the morphosyntatic agreement between verbs and their subjects, and between nouns and their adjectives. However, they also interact with a morpholexical feature called \textit{rationality} -- a feature that is associated with human actors \cite{alkuhlani-habash-2011-corpus}. For example, adjectives modifying rational nouns agree with them in gender and number, while adjectives modifying irrational plural nouns are always feminine and singular.

\paragraph{Orthographic Ambiguity and Noise} In addition to its morphological richness and complexity, Arabic is also orthographically ambiguous as it uses optional diacritics to specify short vowels and consonantal doubling. As these diacritics are optional, Arabic readers deduce the meaning of words based on the sentential context. Since some gender-specific words only differ in diacritics, Arabic orthography makes such  distinctions ambiguous. 
In the context of text generation, this is sometimes a useful feature as it allows the same text to be interpreted differently by the target users. For example, the question 
 \<ما اسمك؟>  \textit{mA Asmk?} `what's your name?' can be diacritized as 
 \<مَا اسْمُكَ؟>  
 \textit{mA  Asmuka?} [2nd.m.sg]
 or
 \<ما اسْمُكِ؟>   
 \textit{mA Asmuki?} [2nd.f.sg].

Finally, it has been shown that unedited MSA text could have a significant percentage ($\sim$23\%) of spelling errors \cite{Zaghouani:2014:large}. The most common errors include Alif-Hamza spelling (\<أ, إ, آ, ا> {\it A, {\AMADDA}, {\AHAMZADN}, {\AHAMZAUP}}), Ya spelling  (\<ى, ي>   {\it y, {\AMAQSURA}}), and the feminine singular suffix Ta-Marbuta (\<ة, ه>  {\it h,~{\TAMARBUTA}}). Therefore, Alif/Ya normalization is a standard processing in Arabic NLP as it reduces some of the noise \cite{Habash:2010:introduction}. This high degree of orthographic ambiguity and noise poses challenges for automatic learning systems due to confusability and data sparsity. 

\section{Extending the Arabic Parallel Gender Corpus}
\label{sec:corpus}

In this section, we describe the selection criteria and the annotation process of the APGC~v2.0. To the best of our knowledge, no such corpus exists for Arabic or any other language.

\begin{table*}[h]
\begin{center}
\setlength{\tabcolsep}{2pt}
\scalebox{0.85}{
\begin{tabular}{|l|r|c|c|r|c}
\cline{1-5}

\multicolumn{1}{|c|}{\bf English} & \multicolumn{1}{c|}{\bf Arabic} & \bf Label & \bf Reinflection Label & \multicolumn{1}{c|}{\bf Reinflection} & \\\cline{1-5}\cline{1-5} 

I wanna thank you & \<أريد أن أشكرك> & BB &  & & (a) \\\cline{1-5}\cline{1-5} 
I have something to say & \<لدي شيء لأقوله> & BN &  & & (b) \\\cline{1-5}\cline{1-5} 
I'm so  \textcolor{blue}{happy} for you & \<من أجلك> \textcolor{blue}{\<سعيدة>} \<أنا> & FB & MB & \<من أجلك> \textcolor{blue}{\<سعيد>} \<أنا> & (c) \\\cline{1-5}
We were  \textcolor{blue}{coming} to see you & \<لرؤيتك> \textcolor{blue}{\<قادمات>} \<نحن> & F!B & M!B & \<لرؤيتك> \textcolor{blue}{\<قادمون>} \<نحن> & (d) \\\cline{1-5}\cline{1-5}
Because I'm your big \textcolor{blue}{brother}  &\<الكبير>  \textcolor{blue}{ \<أخوك> } \<لأنني>  & MB & FB & \<الكبيرة> \textcolor{blue}{\<أختك>}  \<لأنني> & (e) \\\cline{1-5}
We're \textcolor{blue}{ready}  & \textcolor{blue}{\<مستعدون>} \<نحن> & M!B & F!B & \textcolor{blue}{\<مستعدات>} \<نحن> & (f) \\\cline{1-5}\cline{1-5}

I know, \textcolor{red}{babe}  & \textcolor{red}{\<عزيزتي>} \<أعلم ذلك يا> & BF & BM &  \textcolor{red}{\<عزيزي>} \<أعلم ذلك يا> & (g)\\\cline{1-5}
I \textcolor{red}{respect} you [pl]  &   \textcolor{red}{\<أحترمكن>}   \<أنا> & BF! & BM! &  \textcolor{red}{\<أحترمكم>} \<أنا> & (h) \\\cline{1-5}\cline{1-5}
I'm right here \textcolor{red}{dad}  & \textcolor{red}{\<أبي>}  \<أنا هنا يا> & BM & BF & \textcolor{red}{\<أمي>} \<أنا هنا يا> & (i) \\\cline{1-5}
I \textcolor{red}{love} you [pl] so much  & \<كثيرا> \textcolor{red}{\<أحبكم>} & BM! & BF! &  \<كثيرا> \textcolor{red}{\<أحبكن>} & (j) \\\cline{1-5}\cline{1-5}

 &  & & FM & \textcolor{red}{\<ترحل>} \<يجب أن> \textcolor{blue}{\<آسفة>} & (k) \\
I'm \textcolor{blue}{sorry}, you're going to have to \textcolor{red}{leave} & \textcolor{red}{\<ترحل>} \<يجب أن> \textcolor{blue}{\<آسف>} & MM & MF & \textcolor{red}{\<ترحلي>} \<يجب أن> \textcolor{blue}{\<آسف>} & (l) \\
 & & & FF & \textcolor{red}{\<ترحلي>} \<يجب أن> \textcolor{blue}{\<آسفة>} & (m) \\\cline{1-5}

 & &  & MM & \textcolor{red}{\<عزيزي>}  \<للغاية يا>  \textcolor{blue}{\<خائف>}  \<أنا> & (n) \\
 \textcolor{red}{Baby}, I'm so \textcolor{blue}{scared} right now & \textcolor{red}{\<عزيزي>}  \<للغاية يا>  \textcolor{blue}{\<خائفة>} \<أنا> & FM & FF & \textcolor{red}{\<عزيزتي>}  \<للغاية يا>  \textcolor{blue}{\<خائفة>}  \<أنا> & (o) \\
 & & & MF & \textcolor{red}{\<عزيزتي>}  \<للغاية يا>  \textcolor{blue}{\<خائف>}  \<أنا> & (p) \\\cline{1-5}
 
  &  &  & FF & \textcolor{red}{\<أماه>} \<بعودتك يا> \textcolor{blue}{\<سعيدة>} \<أنا> & (q) \\
I'm \textcolor{blue}{glad} you made it home, \textcolor{red}{mom}  & \textcolor{red}{\<أماه>} \<بعودتك يا> \textcolor{blue}{\<سعيد>} \<أنا> & MF & MM & \textcolor{red}{\<أبتاه>}  \<بعودتك يا> \textcolor{blue}{\<سعيد>} \<أنا> & (r) \\
 & & & FM & \textcolor{red}{\<أبتاه>}  \<بعودتك يا> \textcolor{blue}{\<سعيدة>} \<أنا> & (s) \\\cline{1-5}
 
  & & & MF & \textcolor{blue}{\<بالغبي>} \textcolor{red}{\<تناديني>} \<لا> & (t) \\
 Don't \textcolor{red}{call} me a \textcolor{blue}{fool} & \textcolor{blue}{\<بالغبية>} \textcolor{red}{\<تناديني>} \<لا> & FF & FM & \textcolor{blue}{\<بالغبية>} \textcolor{red}{\<تنادني>} \<لا> & (u) \\
 & & & MM & \textcolor{blue}{\<بالغبي>} \textcolor{red}{\<تنادني>} \<لا> & (v) 
 \\\cline{1-5}
 
\end{tabular}
}
\end{center}
\caption{Examples from the Arabic Parallel Gender Corpus v2.0 including the original sentence, its gender label, its reinflection gender label, and its reinflection/rewrite to the opposite grammatical gender where appropriate. First person gendered words are in \textcolor{blue}{blue} and second person gendered words are in \textcolor{red}{red}. The ``!'' in the labels indicate plural forms were used. The two-letter reinflection label specifies gender information of first person (first letter) and second person (second letter). M is Masculine;  F is Feminine; B is invariant; and N is non-existent.}
\label{tab:examples}
\end{table*}

\subsection{Corpus Selection}
As in \newcite{habash-etal-2019-automatic},
we selected the original set of sentences in our corpus from the English-Arabic OpenSubtitles 2018 dataset \cite{lison-tiedemann-2016-opensubtitles2016}, which includes 29.8 million English-Arabic sentence pairs. 
We chose OpenSubtitles because it has parallel sentences in English and because it is full with conversational (first and second person) texts in MSA.
We extracted all the pairs that include first or second person pronouns on the English side: \textit{I, me, my, mine, myself}, and \textit{you, your, yours, yourself}. This selection process identified 13.4 million pairs: 2.8 million (21.1\%) include first and second person pronouns, 5.7 million (42.5\%) include only first person pronouns, and 4.9 million (36.4\%) include only second person pronouns. 
Out of this set, we selected 52,000 English-Arabic pairs to be manually annotated, while maintaining the original first and second person sentences proportions: 10,972 (21.1\%) pairs contain first and second person pronouns on the English side, 22,100 (42.5\%) pairs contain only first person pronouns on the English side, 18,928 (36.4\%) pairs contain only second person pronouns on the English side. To be consistent with APGC~v1.0's preprocessing,
we ran the Arabic sentences through MADAMIRA \cite{Pasha:2014:madamira} to do white-space-and-punctuation tokenization and UTF-8 cleaning.

In addition to the above, we 
decided to re-annotate all of the 11,240 sentences from APGC~v1.0 to include second person references and match our extended guidelines completely. 
In total, this resulted in 63,240 English-Arabic sentence pairs for the next annotation step. 
In the final released corpus, we provide labels indicating the origins of all the sentences.

\begin{table*}[h]
\begin{center}
\setlength{\tabcolsep}{2pt}
\scalebox{0.85}{
\begin{tabular}{|l|rrrrr|c|l}
\cline{1-7}

\multicolumn{1}{|c|}{\bf English} & \multicolumn{5}{c|}{\bf Arabic} & \bf Label \\\cline{1-7}

I wanna thank you & & & \<أشكرك> & \<أن> & \<أريد> &  BB & (a)\\
&  &  &  B & B & B & \\\cline{1-7}

I have something to say & & &  \<لأقوله> &  \<شيء>  & \<لدي> & BN & (b) \\
&  &  &  B & B & B & \\\cline{1-7}

\multirow{4}{*}{I'm so  \textcolor{blue}{happy} for you} & & \<أجلك>  & \<من> & \textcolor{blue}{\<سعيدة>}  & \<أنا> & FB & (c) \\
& & B & B & \textcolor{blue}{1F} & B & \\
&  & \<أجلك>  & \<من> &  \textcolor{blue}{\<سعيد>}  & \<أنا> & MB & (d) \\
& & B & B & \textcolor{blue}{1M} & B & \\\cline{1-7}

\multirow{4}{*}{I know, \textcolor{red}{babe}} & & \textcolor{red}{\<عزيزتي>}  & \<يا> &  \<ذلك>  & \<أعلم> & BF & (e) \\
& & \textcolor{red}{2F} & B & B & B &\\
& & \textcolor{red}{\<عزيزي>}  & \<يا> &  \<ذلك>  & \<أعلم> & BM & (f)\\
& & \textcolor{red}{2M} & B & B & B &\\\cline{1-7}

\multirow{8}{*}{\textcolor{red}{Baby}, I'm so \textcolor{blue}{scared} right now} & \textcolor{red}{\<عزيزي>}  & \<يا> & \<للغاية> &  \textcolor{blue}{\<خائفة>}  & \<أنا> & FM & (g) \\
& \textcolor{red}{2M} & B & B & \textcolor{blue}{1F} & B & \\
& \textcolor{red}{\<عزيزي>}  & \<يا> & \<للغاية> &  \textcolor{blue}{\<خائف>}  & \<أنا> & MM & (h)\\
& \textcolor{red}{2M} & B & B & \textcolor{blue}{1M} & B & \\
& \textcolor{red}{\<عزيزتي>}  & \<يا> & \<للغاية> &  \textcolor{blue}{\<خائفة>}  & \<أنا> & FF & (i) \\
& \textcolor{red}{2F} & B & B & \textcolor{blue}{1F} & B & \\
& \textcolor{red}{\<عزيزتي>}  & \<يا> & \<للغاية> &  \textcolor{blue}{\<خائف>}  & \<أنا> & MF & (j)\\
& \textcolor{red}{2F} & B & B & \textcolor{blue}{1M} & B & \\\cline{1-7}

\end{tabular}
}
\end{center}
\caption{Examples of word-level gender annotation. First person gendered words are in \textcolor{blue}{blue} and second person gendered words are in \textcolor{red}{red}.}
\label{tab:word-level-example}
\end{table*}


\subsection{Corpus Annotation} 
\label{sec:annotation}
We conducted the annotation through a linguistic annotation firm that hired professional linguists to complete the task.\footnote{\url{https://www.ramitechs.com/}.} 
We provided them with the annotation guidelines available in Appendix~\ref{sec:appendix-guidelines}.

\paragraph{Gender Identification} First, the annotators were asked to identify the genders of the first and second person references in each sentence. Then assign to each sentence a two-letter label,  where each letter refers to the gender of the first and second person references, respectively. Each letter in the label can have one of four values: F (feminine), M (masculine), B (invariant/ambiguous), or N (Non-existent). Therefore, each sentence will get a label from one of the 16 different label combinations -- BB, FB, MB, BF, BM, BN, NB, NN, FN, MN, NF, NM, MM, FM, MF, or FF. Additionally, the annotators were asked to identify the dual and plural gendered references. In case they exist, the sub-label corresponding to the gender of the first or second person reference would get an extra mark: ``!'' (e.g., BF!, M!B!, etc.). 

\paragraph{Gender Reinflection/Rewriting}
In the case of an F or M sub-label, the annotators were asked to copy the sentence and modify it to obtain the opposite gender forms.
The modifications are strictly limited to morphological reinflections and word substitutions as was done in \cite{habash-etal-2019-automatic}. Therefore, the total number of words is maintained along with a perfect alignment between each sentence and its parallel opposite gender forms. For example, the sentence in Table~\ref{tab:examples}(c) includes a first person gender reference and is labeled by the annotators as FB, and therefore, the annotators would introduce its gender cognate MB. If the sentence includes both first and second person gender references (MM, FM, MF, or FF), the annotators would then introduce all its possible gender cognates, as in Table~\ref{tab:examples}(k-m) for instance.

In the vast majority of cases, the opposite gender forms of most words end up sharing the same lemma (reinflection), e.g., \<والد> \textit{wAld} `parent/father [M]' and \<والدة> \textit{wAld\TAMARBUTA} `parent/mother [F]'.
However, there are cases where gender-specific words have to be mapped to different lemmas, resulting in a lexical change. For instance, \<أبي> \textit{\AHAMZAUP by} `my dad' and \<أمي>  \textit{\AHAMZAUP my} `my mom' (Table \ref{tab:examples}(i)), or  \<أخوك> \textit{\AHAMZAUP xwk} `your brother' and \<أختك> \textit{\AHAMZAUP xtk} `your sister' (Table~\ref{tab:examples}(e)).\footnote{While technically these are instances of lexical rewriting and not morphological reinflection, we interchangeably refer to the whole process covering both phenomena as reinflection or rewriting.}

Furthermore, the annotators were instructed to avoid any heterocentric assumptions during the annotation. For example, the sentence \<أنت زوجي>
\textit{{\AHAMZAUP}nt ~zwjy} `you are my husband' is labeled as BM (ambiguous first person, masculine second person) and not FM (feminine first person, masculine second person) . The annotators were also instructed to treat all proper names as gender-ambiguous (B), even when they have strong gender-specific associations, and as such are not rewritten. Finally, the annotators were asked to flag bad translations and malformed sentences. 

At the end of the annotation process, we did a quality check on the dataset and fixed some of the annotation errors manually. Most of these errors were either due to malformed Arabic subtitles or misalignment between the parallel sentences. 

\begin{table*}[t!]
\begin{center}
\scalebox{0.7}{

\begin{tabular}{|rr|c|ccc|  ccc  |c|c|c|c|c|rr|}
\multicolumn{2}{c}{} & \multicolumn{4}{c}{\bf(a)} & & & \multicolumn{6}{c}{\bf (b)} & \multicolumn{1}{c}{} \\
\cline{3-6}\cline{10-14}
\multicolumn{2}{c|}{}& \multicolumn{4}{c|}{\bf Original Corpus} & & & & \multicolumn{5}{c|}{\bf Balanced Corpus} & \multicolumn{2}{c}{} \\\cline{1-6}\cline{10-16}

\multicolumn{2}{|c|}{\bf Sentences} & \bf  Label &  \multicolumn{3}{c|}{\bf Reinflection Label}   &  & & &  \bf Input & \bf \bf Target\textsubscript{\textit{MM}} &  \bf Target\textsubscript{\textit{FM}} & \bf \bf Target\textsubscript{\textit{MF}} &  \bf \bf Target\textsubscript{\textit{FF}} & \multicolumn{2}{c|}{\bf Sentences} \\\cline{1-6}\cline{10-16}

    36,980 & 63.7\% & BB &  & & & & & & BB & BB & BB & BB &  BB & 36,980 &  46\%\\\hhline{*{6}{=}*{3}{~}*{7}{=}}
    
    1,123 & 1.9\% & FB &  & MB & & & & & FB & MB &  FB & MB &  FB &  3,063 &  3.8\%\\\cline{1-6}\cline{10-16}
    1,940 & 3.3\% & MB &  & FB & & & & & MB & MB &  FB & MB &  FB &  3,063 & 3.8\%\\\hhline{*{6}{=}*{3}{~}*{7}{=}}

    5,210 & 9\% & BF &  & BM & &  & & & BF & BM &  BM  & BF &  BF & 17,374 & 21.6\%\\\cline{1-6}\cline{10-16}
    12,164 & 21\% & BM &  & BF & &  & & & BM & BM &  BM  & BF &  BF & 17,374 & 21.6\%\\\hhline{*{6}{=}*{3}{~}*{7}{=}}
    
    68 & 0.1\% & FF &  MF & FM & MM & & & & FF & MM &  FM & MF &  FF & 618 & 0.8\%\\\cline{1-6}\cline{10-16}
    135 & 0.2\% &FM &  MM & FF & MF & & & & FM & MM &  FM & MF &  FF &  618 & 0.8\%\\\cline{1-6}\cline{10-16}
    117 & 0.2\% & MF &  FF & MM & FM & & & & MF & MM & FM & MF & FF &   618& 0.8\% \\\cline{1-6}\cline{10-16}
    298 & 0.5\% & MM &  FM & MF & FF & & & & MM & MM & FM & MF & FF &  618 & 0.8\%\\\cline{1-6}\cline{10-16}

    58,035 &  & \multicolumn{4}{c}{} & & &   \multicolumn{6}{c|}{} & 80,326 &  \\\cline{1-2}\cline{15-16}

\end{tabular}
}
\end{center}
\caption{Sentence-level statistics of the original corpus (a) and the balanced corpus (b) with its five versions.}
\label{tab:sentence-level-stats}
\end{table*}

\begin{table*}[t!]
\begin{center}
\scalebox{0.7}{

\begin{tabular}{|rr|c|c|  ccc  |c|c|c|c|c|rr|}
\multicolumn{2}{c}{} & \multicolumn{2}{c}{\bf(a)} & & & \multicolumn{6}{c}{\bf (b)} & \multicolumn{2}{c}{} \\
\cline{3-4}\cline{8-12}
\multicolumn{2}{c|}{} & \multicolumn{2}{c|}{\bf Original Corpus} & & & & \multicolumn{5}{c|}{\bf Balanced Corpus} & \multicolumn{2}{c}{} \\\cline{1-4}\cline{8-14}

\multicolumn{2}{|c|}{\bf Words} &\bf  Label &  \multicolumn{1}{c|}{\bf Reinflection Label}   &  & & &  \bf Input & \bf \bf Target\textsubscript{\textit{MM}} &  \bf Target\textsubscript{\textit{FM}} & \bf \bf Target\textsubscript{\textit{MF}} &  \bf \bf Target\textsubscript{\textit{FF}} & \multicolumn{2}{c|}{\bf Words} \\\cline{1-4}\cline{8-14}

395,658 & 93.5\% & B & & & & & B & B & B & B &  B & 538,733  & 90.3\% \\\hhline{*{4}{=}*{3}{~}*{7}{=}}
1,511 & 0.4\% & 1F & 1M & & & & 1F & 1M & 1F & 1M &  1F & 4,923 &  0.8\% \\\cline{1-4}\cline{8-14}
2,716 & 0.6\% & 1M & 1F & & & & 1M & 1M & 1F & 1M &  1F & 4,923  & 0.8\% \\\hhline{*{4}{=}*{3}{~}*{7}{=}}
6,844  & 1.6\% & 2F & 2M & & & & 2F & 2M & 2M & 2F &  2F & 24,110 &  4\%  \\\cline{1-4}\cline{8-14}
16,525 & 3.9\% & 2M & 2F & & & & 2M & 2M & 2M & 2F &  2F & 24,110  &  4\% \\\cline{1-4}\cline{8-14}
423,254 &  & \multicolumn{10}{c|}{} & 596,799 & \\\cline{1-2}\cline{13-14}

\end{tabular}
}
\end{center}
\caption{Word-level statistics of the original corpus (a) and the balanced corpus (b) with its five versions.}
\label{tab:word-level-stats}
\end{table*}






\subsection{Automatic Word-Level Annotations}
Given that the annotators were only allowed to perform grammatical inflections and word substitutions when introducing the opposite gender forms of a particular sentence, all sentences and their parallels are perfectly aligned at the word level. This allowed us to obtain word-level gender annotations automatically as a byproduct. To do this, we look at the original sentence and all of its parallel forms. If the word is the same across all the parallel versions of a sentence, then we label it as B. Otherwise, we assign the word a label based on its sentence-level gender label. For example, in Table~\ref{tab:word-level-example}(g-j), the word \<أنا> \textit{\AHAMZAUP na} `I' is the same across all four parallel versions of the sentence and thus labeled as B. In contrast, the words \<خائفة> 
\textit{xa\YHAMZA f\TAMARBUTA} `scared [F]' and \<خائف> \textit{xa\YHAMZA f} `scared [M]' change across the parallel versions. By looking at the sentence-level labels of the four parallel forms, we can deduce that the word \<خائفة> \textit{xa\YHAMZA f\TAMARBUTA} is first-person feminine and will be labeled as 1F, and that the word \<خائف> \textit{xa\YHAMZA f} is first-person masculine and will be labeled as 1M. Similarly, we determine that the words \<عزيزي> \textit{\AYN zyzy} `baby/dear [M]' \<عزيزتي> \textit{{\AYN}zyzty} `baby/dear [F]' are second-person masculine and second-person feminine and are labeled as 2M and 2F, respectively. All words belonging to sentences that do not have any gender cognates (BB, BN, NB, etc. cases) as in Table~\ref{tab:word-level-example}(a and b) are labeled as B. Therefore, each word can have one of following possible labels: B, 1F, 1M, 2F, 2M. We also mark the dual/plural words by adding ``!'' to their corresponding labels. 

\section{Corpus Overview and Statistics}
\label{sec:overview-and-analysis}
\subsection{The Original Corpus}

After the annotation, 8.2\% of the sentences (5,205) were eliminated due to malformed Arabic and annotation errors. This resulted in 58,035 (423,254 words) sentences, constituting our Original Corpus.
We created a condensed version of the annotations for this corpus by mapping the N (non-existent) sub-labels to B (invariant/ambiguous) and removing the dual/plural marks (``!'') from the labels across all the sentences.

\paragraph{Corpus Statistics}
Table~\ref{tab:sentence-level-stats}(a) includes the statistics about the Original Corpus. Out of all sentences, 36,980 (63.7\%) are labeled as BB. There are 17,374 (30\%) sentences that include only second-person gendered references (BF and BM). This is five times more than sentences with only first-person gendered references (FB and MB), which accounts for 5.3\% (3,063 sentences) of all sentences. Moreover, the number of sentences including first or second person masculine references is more than the ones including feminine references (12,164 BM vs 5,210 BF, and 1,940 MB vs 1,123 FB). There are 618 (1.1\%) sentences that have both first and second gendered references. All of the sentences which have first or second (or both) person gendered references are rewritten to introduce their opposite gender forms. This resulted in 21,055 manually added sentences (162,055 words). The word-level statistics of our Original Corpus are shown in Table~\ref{tab:word-level-stats}(a). Among the newly added sentences, about 17\% (27,596) of the words are gender-specific, constituting around 6.5\% of all the words in the corpus.

Table~\ref{tab:sentence-level-stats-fg}(a) and Table~\ref{tab:word-level-stats-fg}(a) in Appendix~\ref{sec:appendix-stats} present statistics on the non-condensed annotations of the Original Corpus at the sentence and word levels, respectively.

\paragraph{Morphological Reinflection vs Lexical Rewriting}
To quantify the proportions of the morphological reinflections and lexical changes introduced as part of the manual annotation process (\S\ref{sec:annotation}), we analyzed the gender-specific words across all parallel sentences using the CALIMA$_{Star}$ Arabic morphological analyzer \cite{Taji:2018:arabic-morphological} included in the CAMeL~Tools toolkit \cite{obeid-etal-2020-camel}. We consider the manually introduced gender cognate of a specific word to be its reinflection, if both words share at least one lemma. If no lemmas are shared, then the gender cognate is a result of a lexical change. If the word or its gender cognate does not get recognized by the morphological analyzer, we look at them manually. Out of the 27,596 newly introduced gender specific words, 26,728 (96.9\%) resulted from morphological reinflection, whereas 868 words (3.1\%) resulted from lexical rewriting.

\subsection{The Balanced Corpus}

Similarly to \newcite{habash-etal-2019-automatic}, to ensure equal gender representation in our dataset, we force balance the corpus by adding the manually rewritten sentences to the Original Corpus and using their original forms as their rewritten forms. This constitutes our Balanced Corpus.


\paragraph{Corpus Statistics}
The sentence-level  statistics of the Balanced Corpus are presented in
Table~\ref{tab:sentence-level-stats}(b). 
This corpus has 80,326 sentences in total. Out of all sentences, 46\% (36,980) are marked as BB, whereas sentences with gendered references constituted 54\% (43,346 sentences). We introduce five versions of the Balanced Corpus: Input, Target\textsubscript{\textit{MM}}, Target\textsubscript{\textit{FM}}, Target\textsubscript{\textit{MF}}, and Target\textsubscript{\textit{FF}}. The balanced Input Corpus, includes all the sentences from the Original Corpus in addition to their rewritten forms. 
The Target\textsubscript{\textit{MM}} corpus is the masculine-only corpus and it includes sentences that are either invariant/ambiguous or have a first or second person (or both) masculine references. Therefore, it only contains BB, MB, BM, and MM sentences. 
The Target\textsubscript{\textit{MF}} corpus is the masculine-feminine corpus and it contains sentences that are either invariant/ambiguous or have first person masculine references, second person feminine references, or first person masculine and second person feminine references (i.e., BB, MB, BF, and MF sentences). 
The Target\textsubscript{\textit{FM}} corpus is the feminine-masculine corpus and it contains BB, FB, BM, and FM sentences. 
Finally, the Target\textsubscript{\textit{FF}} corpus is the feminine-only corpus and it contains BB, FB, BF, and FF sentences. 
All five corpora have the same number of sentences, words, and gendered-specific words. The word-level statistics of the Balanced Corpus are shown in Table~\ref{tab:word-level-stats}(b). Table~\ref{tab:sentence-level-stats-fg}(b) and Table~\ref{tab:word-level-stats-fg}(b) in Appendix~\ref{sec:appendix-stats} present statistics on the non-condensed annotations of the Balanced Corpus at the sentence and word levels, respectively.

\paragraph{Corpus Splits}
To aid reproducibility when using APGC~v2.0 for various research experiments, we provide train, development, and test splits for all five balanced corpora. Following \newcite{habash-etal-2019-automatic}, all five corpora were divided randomly as follows: training ({\sc TRAIN}: 70\% or 57,603 sentences), development ({\sc DEV}: 10\% or 6,647 sentences) and testing  ({\sc TEST}: 20\% or 16,076 sentences). We made sure that the splits are balanced and all parallel versions of the sentences are in the same split.  




\section{Revisiting the Motivation: Quantifying Bias in Gender-Unaware Machine Translation}
\label{sec:machine-translation}
\begin{table*}[ht!]
\centering
\setlength{\tabcolsep}{3pt} \scalebox{1}{
\begin{tabular}{|llll|c|c|c|c|c||c|}
\cline{1-10}
\multicolumn{4}{|p{3cm}|}{\bf Selected Sentences\textsubscript{\textit{ar}}} &\multicolumn{1}{c|}{\multirow{2}{*}{\bf Count}} &\multicolumn{1}{c|}{\multirow{2}{*}{\bf Target\textsubscript{\textit{MM}}}}  & \multicolumn{1}{c|}{\multirow{2}{*}{\bf Target\textsubscript{\textit{FM}}}} & \multicolumn{1}{c|}{\multirow{2}{*}{\bf Target\textsubscript{\textit{MF}}}}& \multicolumn{1}{c||}{\multirow{2}{*}{\bf Target\textsubscript{\textit{FF}}}} & \multicolumn{1}{c|}{\multirow{2}{*}{\bf Multi-Reference}} \\\hline\hline
ALL &  & & & 80,326 & 13.5 & 13.1 & 11.4 & 11.0 & 13.6 \\\hline\hline
BB &  & & & 36,980 & 14.0 & 14.0 & 14.0 & 14.0 & 14.0 \\\hline
\multicolumn{4}{|l|}{ALL - BB} & 43,346 & 13.1 & 12.4 &  9.3 & 8.6 & 13.4 \\\hline\hline
BM & BF &  &  & 34,748 & 13.1 & 13.1 & 8.6 & 8.6 & 13.3 \\\hline
MB & FB & & & 6,126 & 12.9 & 9.6 & 12.9 & 9.6  & 13.6\\\hline
MM & FM & MF & FF & 2,472 & 12.9 & 9.5 & 9.5 & 6.7 & 13.5  \\\hline
\end{tabular}
}
\caption{BLEU results (all Alif/Ya/Ta-Marbuta normalized) of the English-Arabic Google Translate output for the balanced input corpus against the four balanced target corpora.}
\label{tab:bleu-google}
\end{table*}

\begin{table*}[t!]
\centering
\setlength{\tabcolsep}{3pt} \scalebox{1}{
\begin{tabular}{|l|l|c|c|c|c|c||c|}
\cline{1-8}
\multicolumn{1}{|p{2cm}|}{\bf Selected Sentences\textsubscript{\textit{ar}}} & \multicolumn{1}{|p{2cm}|}{\bf Selected Sentences\textsubscript{\textit{en}}} &\multicolumn{1}{c|}{\multirow{2}{*}{\bf Count}} & \multicolumn{1}{c|}{\multirow{2}{*}{\bf Target\textsubscript{\textit{MM}}}}  & \multicolumn{1}{c|}{\multirow{2}{*}{\bf Target\textsubscript{\textit{FM}}}} & \multicolumn{1}{c|}{\multirow{2}{*}{\bf Target\textsubscript{\textit{MF}}}}& \multicolumn{1}{c||}{\multirow{2}{*}{\bf Target\textsubscript{\textit{FF}}}} & \multicolumn{1}{c|}{\multirow{2}{*}{\bf Multi-Reference}} \\\hline\hline
ALL - BB  & ALL$_{en}$ &  43,346 & 13.1  &  12.4 &  9.3 & 8.6 & 13.4 \\\hline\hline
ALL - BB  & B$_{en}$ & 39,484  & 13.1 & 12.4 & 9.2 & 8.5 & 13.3 \\\hline\hline
ALL - BB  & M$_{en}$ & 2,606 & 14.1 &  13.0 & 9.6 & 8.5 & 14.2 \\\hline\hline
ALL - BB  & F$_{en}$ & 1,256 & 10.8  & 11.1 & 10.0 & 10.4 & 13.2 \\\hline
\end{tabular}
}
\caption{BLEU results (all Alif/Ya/Ta-Marbuta normalized) of the English-Arabic Google Translate 
output for the English sentences corresponding to gender specific (ALL - BB) Arabic sentences as in Table~\ref{tab:bleu-google}.}

\label{tab:bleu-google-en}
\end{table*}

The efforts to develop APGC~v1.0 and APGC~v2.0 were motivated by the observation of common gender bias in gender-unaware NLP systems targeting morphologically rich languages, specifically Arabic in our case.
In this section, we revisit this motivation and 
use our newly created corpus to quantify and detect gender bias in machine translation. We translated the English side of the Input balanced corpus to Arabic using the Google Translate API.\footnote{\url{https://cloud.google.com/translate} on September \nth{15}, 2021} 
We chose to use Google Translate because of its popularity, but these experiments can be easily done on any machine translation output.\footnote{It should be noted with admiration that the Google Translate team has done a lot of work on the front of fighting gender bias, e.g., generating multiple gendered translation for some language pairs \cite{Johnson:2020:Approach}. To date Arabic is not one of these languages.}
We include Google Translate's outputs in the release of our corpus to encourage research and development on corrective post-editing.

In Tables~\ref{tab:bleu-google}~and~\ref{tab:bleu-google-en}, we present the English-Arabic Google Translate results 
in terms of BLEU~\cite{Papineni:2002:bleu} using the latest version of SacreBLEU~\cite{post-2018-call}. All the results are reported in an orthographically normalized space for Alif, Ya, and Ta-Marbuta~\cite{Habash:2010:introduction}. 
We evaluate Google Translate's output against all four balanced target corpora (i.e., Target\textsubscript{\textit{MM}}, Target\textsubscript{\textit{FM}}, Target\textsubscript{\textit{MF}}, Target\textsubscript{\textit{FF}}) separately as well as in a multi-reference setting.
The presented results are organized around different subsets of the Balanced Corpus to allow us to determine the effect of different gender specificity factors in Arabic and English on the results.


\subsection{Overall Results}
To start off, looking at the BLEU scores of all sentences ({\sc ALL}) in Table~\ref{tab:bleu-google}, we notice that the score against the Target\textsubscript{\textit{MM}} corpus is higher than the score against Target\textsubscript{\textit{FF}} (by 2.5 BLEU absolute). Moreover, we notice that the multi-reference BLEU score is a little higher than the score of Target\textsubscript{\textit{MM}} (0.1 BLEU). This indicates that scores from the evaluation against Target\textsubscript{\textit{FM}}, Target\textsubscript{\textit{MF}}, and Target\textsubscript{\textit{FF}} are contributing to the overall increase in the multi-reference evaluation but not that much. From these basic results, we observe  that every time an M participant is switched to F, the BLEU scores drop. This strongly suggests that the machine translation output is biased towards masculine grammatical gender preferences.

\subsection{Results on Arabic Gender Specific Subsets}
The remainder of Table~\ref{tab:bleu-google} presents the results organized by Arabic gender-specificity factors.
The Arabic invariant/ambiguous ({\sc BB}) sentences have the same BLEU scores in all conditions because they do not vary across the different Target references. 
When we compare the BLEU scores of BB sentences with gender specific sentences ({\sc ALL~-~BB}), we notice a 0.6 drop in the multi-reference evaluation. This indicates that sentences with gender specific words are harder for Google Translate than gender invariant/ambiguous sentences. 

The drop in BLEU scores from BB to {\sc ALL~-~BB} for Target\textsubscript{\textit{FF}} is 5.4 BLEU, which is six times the corresponding drop for Target\textsubscript{\textit{MM}}.
Also, the difference between the Target\textsubscript{\textit{MM}} and Target\textsubscript{\textit{FF}} BLEU scores for ({\sc ALL~-~BB}) is almost double the difference for {\sc ALL} (4.5 vs 2.5).
By grouping the Arabic gender-marked sentences ({\sc ALL~-~BB}) based on the variation of the first and second person gendered references (i.e., BM, BF, MB, FB, MM, FM, MF, and FF), we again observe that every time an M participant is switched to F, the BLEU scores drop. In the most extreme case of gender specific references in both first and second person (last row in Table~\ref{tab:bleu-google}), the difference between Target\textsubscript{\textit{MM}} and Target\textsubscript{\textit{FF}} in BLEU scores is 6.2.
Therefore, we can deduce that Google Translate's Arabic outputs are biased against feminine target users compared to masculine users.

\subsection{Results on English Gender Specific Subsets}
While our evaluation setup assumes that Google's Arabic translations could have come from translating sentences in any language into Arabic, we acknowledge that some of the English sentences may be gender specific and the bias we are observing might be caused by such sentences. In this section, we delve into studying the bias in the Arabic translations of gender specific English sentences.

Gender in English is usually expressed referentially through third person pronouns (he, she, they) or lexically using gender specific nouns (e.g., mother, son, etc.)~\cite{cao-daume-iii-2020-toward}. As our parallel gender corpus was only annotated for first and second person gendered Arabic references, we only focus on English sentences that contain gender specific nouns.
We focus on the OpenSubtitles 2018 English sentences corresponding to the gender specific ({\sc ALL - BB}) Arabic sentences in our Balanced Corpus (43,346 sentences). We tokenized each English sentence and obtained the part-of-speech tags of its tokens using spaCy \cite{SpaCy:2017:spacy}. Out of 43,346 sentences, 24,350 contained at least one noun. We annotated these English nouns' lemmas (4,138 unique lemmas) manually as either B$_{en}$ (ambiguous, e.g., {\it teacher, scientist, prostitute}), M$_{en}$ (masculine, e.g., {\it father, policeman, brother}), or F$_{en}$ (feminine, e.g., {\it mother, queen, waitress}). 
Out of the 4,138 lemmas, 97.4\% (4,032) were labeled as B$_{en}$, 1.4\% (60) were labeled as M$_{en}$, and 1.1\% (46) were labeled as F$_{en}$. 
The list of all the English gender-specific noun lemmas is in Appendix~\ref{sec:english-gendered-nouns}.

Using the annotated English nouns, we labeled the English sentences based on the gender of their nouns as follows: if a sentence has only F$_{en}$ or (B$_{en}$ and F$_{en}$) nouns, it is labeled as F$_{en}$. If a sentence has only M$_{en}$ or (B$_{en}$ and M$_{en}$) nouns, it is labeled as M$_{en}$. All other sentences are labeled as B$_{en}$.
This is clearly a rough approach since we do not label the sentences by the first and second person gender reference.
This resulted in 39,484 (91.1\%) B$_{en}$ sentences,  2,606 (6\%) M$_{en}$ sentences, and 1,256 (2.9\%) F$_{en}$ sentences. 
We include the English sentence labels in our released corpus to support further research on this topic.
The results on these subsets are in Table~\ref{tab:bleu-google-en}, which presents the BLEU scores of Google's Arabic translations against our four balanced Target corpora (including multi-reference).

Examining the BLEU scores of the B$_{en}$ sentences, we notice that they are almost identical to the BLEU scores of the ALL$_{en}$ sentences (i.e. ALL~-~BB), which is expected as they are the majority subset. Moreover, although all of the English sentences in this subset  did not have any gender specific nouns, the BLEU score of Google Translate's output against the Target\textsubscript{\textit{\textit{MM}}} sentences is significantly higher than score against the Target\textsubscript{\textit{\textit{FF}}} sentences (4.6 BLEU). This highlights the bias of Google Translate towards masculine users compared to their feminine counterparts when English sentences are invariant or ambiguous gender-wise.

When we consider the BLEU scores of the M$_{en}$ sentences, we observe an expected increase in the scores across the all target corpora with M targets.
As for the F$_{en}$ sentences, we also observe an expected increase for Target\textsubscript{\textit{FF}} paired with a decrease for Target\textsubscript{\textit{MM}}. But even here, the Target\textsubscript{\textit{MM}} BLEU score is still slightly higher that the Target\textsubscript{\textit{FF}} BLEU score.
%



\section{Conclusion and Future Work}
\label{sec:conclusion}
We presented APGC~v2.0, a new Arabic parallel corpus for gender identification and rewriting in contexts involving one or two target users (I and/or You) with independent grammatical gender preferences.  We provided a detailed description of the selection and annotation process we did to create such a corpus. Furthermore, we showed that our corpus can be used to study and quantify the degree and type of gender biases and stereotypes that are embedded in and amplified by one of state-of-the-art commercial machine translation systems.

In future work, we plan to extend our corpus to other languages and dialectal varieties. By building our corpus and making it publicly available, we hope to encourage research on  gender identification, controlled generation, and post-editing rewrite systems that could be used to personalize NLP applications based on their end users' preferences. 

\label{sec:conclusion}

\section*{Acknowledgements}
We thank Ramitechs for their help in the annotation process. We would like to thank Go Inoue and Salam Khalifa for the helpful and insightful conversations.

\section*{Ethical Considerations}
Our dataset is intended to aid the development of gender identification, controlled generation, and post-editing rewrite systems that could be used to personalize NLP applications and provide users with the correct outputs based on their grammatical gender preferences.
We acknowledge that by limiting the choice of gender expression to the grammatical gender choices
in Arabic, we exclude other alternatives such as non-binary gender or no-gender expressions.
We are not aware of any sociolinguistics published research that discusses such alternatives for Arabic, although there are growing grassroots efforts, e.g., the Ebdal Project for Queer Language and Translation.\footnote{\url{https://www.facebook.com/EbdalProject/}} 

\bibliography{anthology,camel-bib-v2,extra}
\bibliographystyle{acl_natbib}
\clearpage
\appendix

\section{Annotation Guidelines}
\label{sec:appendix-guidelines}


# A  Annotation Guidelines

## Gender Identification and Rewriting

**Input**:
The input has three columns: **SentenceID**, **English** and **Arabic**.
These columns should not change throughout the annotation process.

**Output**:
The output has two columns:
**SentenceType**, and **RewrittenSentence**

The value of **SentenceType** should be a two-letter code, where the first code refers to the status of the 1st person reference in the sentence, and the second code refers to the status of the 2nd person reference in the sentence.
The status should be one of four values:
- M: Masculine
- F: Feminine
- B: Ambiguous
- N: does not exist

Here are all the possible combinations with examples:

| Code | Meaning | Example |
|---|---|---|
| M-M | The sentence has a masculine 1st person reference and a masculine 2nd person reference | أنا صبور مثلك يا صديقي |
| M-B | The sentence has a masculine 1st person reference and an ambiguous 2nd person reference | أنا صبور مثلك |
| F-B | The sentence has a feminine 1st person reference and an ambiguous 2nd person reference | مرحبا، أنا جارتك الجديدة |
| B-M | The sentence has an ambiguous 1st person reference and a masculine 2nd person reference | أنا من تنتظره |
| B-F | The sentence has an ambiguous 1st person reference and a feminine 2nd person reference | أحبك يا صديقتي |
| B-B | The sentence has ambiguous 1st person and 2nd person references | أحبك انت |
| M-N | The sentence has a masculine 1st person reference but NO 2nd person reference | أنا قوي |
| F-N | The sentence has a feminine 1st person reference but NO 2nd person reference | أنا سعيدة جدا |
| N-M | The sentence has a masculine 2nd person reference but NO 1st person reference | قف ! |
| N-F | The sentence has a feminine 2nd person reference but NO 1st person reference | انت تكتبين |
| B-N | The sentence has an ambiguous 1st person reference but NO 2nd person reference | حسناً. أنا هنا |
| N-B | The sentence has an ambiguous 2nd person reference but NO 1st person reference | الجو ممتع في بلدك |
| N-N | The sentence has no reference for 1st nor 2nd person | الجو ممتع |

The **RewrittenSentence** column should have the Arabic sentence after converting the masculine references to feminine and the feminine references to masculine.
Here are the changes the sentences above should receive (the comment column does not exist in the annotations but it is just to let you understand the task).

| Code | Arabic | RewrittenSentence | Comment |
|---|---|---|---|
| M-M | أنا صبور مثلك يا صديقي | أنا صبورة مثلك يا صديقي | 1st person changes to go from M-M to F-M |
| M-M | أنا صبور مثلك يا صديقي | أنا صبور مثلك يا صديقتي | 2nd person changes to go from M-M to M-F |
| F-M | أنا صبورة مثلك يا صديقي | أنا صبورة مثلك يا صديقتي | 2nd person changes to go from F-M to F-F |
| M-B | أنا صبور مثلك | أنا صبورة مثلك | |
| F-B | مرحبا، أنا جارتك الجديدة | مرحبا، أنا جارك الجديد | |
| B-M | أنا من تنتظره | أنا من تنتظريه | |
| B-F | أحبك يا صديقتي | أحبك يا صديقي | |
| B-B | أحبك انت | | No change is needed. Leave **RewrittenSentence** empty |
| M-N | أنا قوي | أنا قوية | |
| F-N | أنا سعيدة جدا | أنا سعيد جدا | |
| N-M | قف ! | قفي ! | |
| N-F | انت تكتبين | انت تكتب | |
| B-N | حسناً. أنا هنا | | No change is needed. Leave **RewrittenSentence** empty |
| N-B | الجو ممتع في بلدك | | No change is needed. Leave **RewrittenSentence** empty |
| N-N | الجو ممتع | | No change is needed. Leave **RewrittenSentence** empty |

**VERY IMPORTANT**

**What should we do with 3rd person references?**
Nothing. This task is only about 1st and 2nd person references.
For example: هذا هو الرجل الشجاع is an N-N case.

**How to handle dual and plural cases?**
We do not expect the dual and plural to happen much, but when they do add a "!" to the label (e.g., M!, F! and B!). For example:
لقد سمعتكن تتكلمن عنها should be tagged as B-F!
لقد سمعتك تتكلمين عن أخلاقنا should be tagged as B!-F

**How to handle proper nouns?**
Handle proper nouns as ambiguous. Do not make any assumptions about proper nouns.
For examples: مرحبا، انا ماري should be tagged as B-N
However, مرحبا، انا ماري جارتك الجديدة should be tagged as F-B because of the word الجديدة

**What do to with writing mistakes?**
Do NOT correct any mistakes of any kind.
Do NOT correct even hamzas, Ta' marboota, Alif maqsoura or misplaced punctuation.
Do not add any diacritization. Notice that أحبك انت is a B-B case as diacritization is absent.
If the Arabic sentence has major mistakes (ex: أعرف، طفلرضيع ، أوه ، نعم), just make the sentence type B-S and leave **RewrittenSentence** empty.

**What if the English column does not match the translation in the Arabic column?**
Ignore the English column, do not worry about it. Also, do not use the English column to judge the gender or any other information.
Just use the English column if you need to understand an Arabic sentence that is not clear for some reason.
For example: We know that the word ذهبت in ذهبت إلى هناك refers to 1S not 2MS because the English translation is "I went" not "You went".

**Any hard cases we can encounter?**
Yes, but this is infrequent.
For example: Sometimes you need to change the lemma itself for the gender conversion. Examples:

انا رجل ← انا امرأة
أحبك يا أمي ← أحبك يا أبي
نعم أنا هو ← نعم أنا هي (note that although هي is 3rd person, but we need to change it in order to adjust the 1st person reference).

However, if you encounter any HARD case that you cannot figure out (e.g., انا حامل), just write the word HARD in the **RewrittenSentence** column.



\onecolumn
\section{Fine-Grained Corpus Statistics}
\label{sec:appendix-stats}

\begin{table*}[htb!]
\begin{center}
\scalebox{0.7}{

\begin{tabular}{|rr|c|ccc|  ccc  |c|c|c|c|c|rr|}
\multicolumn{2}{c}{} & \multicolumn{4}{c}{\bf(a)} & & & \multicolumn{6}{c}{\bf (b)} & \multicolumn{2}{c}{} \\
\cline{3-6}\cline{10-14}
\multicolumn{2}{c|}{} & \multicolumn{4}{c|}{\bf Original Corpus} & & & & \multicolumn{5}{c|}{\bf Balanced Corpus} & \multicolumn{2}{c}{} \\\cline{1-6}\cline{10-16}

\multicolumn{2}{|c|}{\bf Sentences} &\bf  Label &  \multicolumn{3}{c|}{\bf Reinflection Label}   &  & & &  \bf Input & \bf \bf Target\textsubscript{\textit{MM}} &  \bf Target\textsubscript{\textit{FM}} & \bf \bf Target\textsubscript{\textit{MF}} &  \bf \bf Target\textsubscript{\textit{FF}} & \multicolumn{2}{c|}{\bf Sentences} \\\cline{1-6}\cline{10-16}

    6,156 & 10.6\% & BB &  & & & & & & BB & BB & BB & BB &  BB & 6,156 & 7.7\% \\\cline{1-6}\cline{10-16}
    601 & 1.0\%  & B!B & & & & & & & B!B & B!B & B!B & B!B &  B!B & 601 & 0.7\% \\\cline{1-6}\cline{10-16}
    148 &  0.3\%  & BB! && & & & & & BB! & BB! & BB! & BB! &  BB! & 148 & 0.2\% \\\cline{1-6}\cline{10-16}
    20 & 0.03\% & B!B! &  & & & & & & B!B! & B!B! & B!B! & B!B! &  B!B! & 20 & 0.02\% \\\cline{1-6}\cline{10-16}
    22,379 & 38.6\% & BN &  & & & & & & BN & BN & BN & BN &  BN & 22,379 & 27.9\% \\\cline{1-6}\cline{10-16}
    457 & 0.8\% & B!N &   & & & & & & B!N & B!N & B!N & B!N &  B!N & 457 & 0.6\% \\\cline{1-6}\cline{10-16}
    5,329 & 9.2\% &NB &  & & & & & & NB & NB & NB & NB &  NB & 5,329 & 6.6\% \\\cline{1-6}\cline{10-16}
    189 & 0.3\% & NB! &  & & & & & & NB! & NB! & NB! & NB! &  NB! & 189 & 0.2\% \\\cline{1-6}\cline{10-16}
    1,701 & 2.9\%  & NN & & & & & & & NN & NN & NN & NN &  NN & 1,701 & 2.1\% \\\hhline{*{6}{=}*{3}{~}*{7}{=}}
 
    254 & 0.4\% & FB & &   MB & & & & & FB & MB &  FB & MB &  FB &  657 & 0.8\% \\\cline{1-6}\cline{10-16}
    403 & 0.7\% & MB & &  FB & & & & & MB & MB &  FB & MB &  FB &  657 & 0.8\% \\\cline{1-6}\cline{10-16}
    2 & 0.003\% & FB! &  &  MB! & & & & & FB! & MB! &  FB! & MB! &  FB! &  9 & 0.01\% \\\cline{1-6}\cline{10-16}
    7 & 0.01 \% & MB! &  &  FB! & & & & & MB! & MB! &  FB! & MB! &  FB! &  9 & 0.01\% \\\cline{1-6}\cline{10-16} 
    0 & 0\% & F!B &  &  M!B & & & & & F!B & M!B &  F!B & M!B &  F!B &  15 & 0.02\% \\\cline{1-6}\cline{10-16} 
    15 & 0.03\% & M!B & &  F!B & & & & & M!B & M!B &  F!B & M!B &  F!B &  15 & 0.02\% \\\cline{1-6}\cline{10-16}
    867 &1.5\% & FN & &  MN & & & & & FN & MN &  FN & MN &  FN &  2,377 & 3.0\% \\\cline{1-6}\cline{10-16}
    1,510 & 2.6\% & MN & & FN & & & & & MN & MN &  FN & MN &  FN &  2,377  & 3.0\% \\\cline{1-6}\cline{10-16}
    0 & 0\% & F!N &  &  M!N & & & & & F!N & M!N &  F!N & M!N &  F!N &  5 & 0.01\% \\\cline{1-6}\cline{10-16}
    5 & 0.01\% & M!N & & F!N & & & & & M!N & M!N &  F!N & M!N &  F!N &  5 & 0.01\% \\\hhline{*{6}{=}*{3}{~}*{7}{=}}

    2,562 & 4.4\% & BF &  & BM & &  & & & BF & BM &  BM  & BF &  BF & 7,733 & 9.6\% \\\cline{1-6}\cline{10-16}
    5,171 & 8.9\% & BM &  & BF & &  & & & BM & BM &  BM  & BF &  BF & 7,733 & 9.6\% \\\cline{1-6}\cline{10-16}
    199 & 0.3\% & B!F &  & B!M & &  & & & B!F & B!M &  B!M  & B!F &  B!F & 632 & 0.8\% \\\cline{1-6}\cline{10-16}
    433 & 0.7\% & B!M &  & B!F & &  & & & B!M & B!M &  B!M  & B!F &  B!F & 632 & 0.8\% \\\cline{1-6}\cline{10-16}
    26 & 0.04\% & BF! &  & BM! & &  & & & BF! & BM! &  BM!  & BF! &  BF! & 877 & 1.1\% \\\cline{1-6}\cline{10-16} 
    851 & 1.5\% & BM! &  & BF! & &  & & & BM! & BM! &  BM!  & BF! &  BF! & 877 & 1.1\% \\\cline{1-6}\cline{10-16}
    1 & 0.002\% & B!F! &  & B!M! & &  & & & B!F! & B!M! &  B!M!  & B!F! &  B!F! & 126 & 0.2\% \\\cline{1-6}\cline{10-16} 
    125 & 0.2\% & B!M! &  & B!F! & &  & & & B!M! & B!M! &  B!M!  & B!F! &  B!F! & 126  & 0.2\% \\\cline{1-6}\cline{10-16}
    2,391 & 4.1\% & NF &  & NM & &  & & & NF & NM &  NM  & NF &  NF & 7,295 & 9.1\% \\\cline{1-6}\cline{10-16}
    4,904 & 8.5\% & NM &  & NF & &  & & & NM & NM &  NM  & NF &  NF & 7,295 & 9.1\% \\\cline{1-6}\cline{10-16} 
    31 & 0.1\% & NF! &  & NM! & &  & & & NF! & NM! &  NM!  & NF! &  NF! & 711 & 0.9\% \\\cline{1-6}\cline{10-16}
    680 & 1.2\% & NM! &  & NF! & &  & & & NM! & NM! &  NM!  & NF! &  NF! & 711 & 0.9\% \\\hhline{*{6}{=}*{3}{~}*{7}{=}}

    64 & 0.1\% & FF &  MF & FM & MM & & & & FF & MM &  FM & MF &  FF & 531 & 0.7\% \\\cline{1-6}\cline{10-16}
    110 & 0.2\% & FM &  MM & FF & MF & & & & FM & MM &  FM & MF &  FF &  531 &0.7\% \\\cline{1-6}\cline{10-16}
    115 & 0.2\% & MF &  FF & MM & FM & & & & MF & MM & FM & MF & FF &   531 & 0.7\% \\\cline{1-6}\cline{10-16}
    242 & 0.4\% & MM &  FM & MF & FF & & & & MM & MM & FM & MF & FF &  531 & 0.7\% \\\cline{1-6}\cline{10-16}
    
    0 & 0\% & F!F &  M!F & F!M & M!M & & & & F!F & M!M & F!M & M!F & F!F & 12 &0.01\%  \\\cline{1-6}\cline{10-16}
    0 & 0\% & F!M &  M!M & F!F & M!F & & & & F!M & M!M & F!M & M!F & F!F &  12 &0.01\%  \\\cline{1-6}\cline{10-16}
    2 & 0.003\% & M!F &  F!F & M!M & F!M & & & & M!F & M!M & F!M & M!F & F!F &  12 & 0.01\%  \\\cline{1-6}\cline{10-16} 
    10 & 0.02\% & M!M &  F!M & M!F & F!F & & & & M!M & M!M & F!M & M!F & F!F & 12  &  0.01\% \\\cline{1-6}\cline{10-16}

    4 & 0.01\% & FF! &  MF! & FM! & MM! & & & & F!F & MM! &  FM! & MF! &  FF! & 72 & 0.1\% \\\cline{1-6}\cline{10-16}
    25 & 0.04\% & FM! &  MM! & FF! & MF! & & & & FM! & MM! &  FM! & MF! &  FF! & 72& 0.1\% \\\cline{1-6}\cline{10-16}
    0 & 0\% & MF! &  FF! & MM! & FM! & & & & MF! & MM! & FM! & MF! & FF! &   72 & 0.1\% \\\cline{1-6}\cline{10-16}
    43 & 0.1\%  & MM! &  FM! & MF! & FF! & & & & MM! & MM! & FM! & MF! & FF! &  72 & 0.1\% \\\cline{1-6}\cline{10-16}

    0 & 0\% & F!F! &  M!F! & F!M! & M!M! & & & & F!F! & M!M! & F!M! & M!F! & F!F! &  3 & 0.004\% \\\cline{1-6}\cline{10-16} 
    0 & 0\% & F!M! &  M!M! & F!F! & M!F! & & & & F!M! & M!M! & F!M! & M!F! & F!F! &  3 & 0.004\% \\\cline{1-6}\cline{10-16}
    0 & 0\% & M!F! &  F!F! & M!M! & F!M! & & & & M!F! & M!M! & F!M! & M!F! & F!F! &  3 & 0.004\%  \\\cline{1-6}\cline{10-16}
    3 & 0.01\% & M!M! &  F!M! & M!F! & F!F! & & & & M!M! & M!M! & F!M! & M!F! & F!F! &  3  & 0.004\% \\\cline{1-6}\cline{10-16}

    58,035   & & \multicolumn{4}{c}{} & & &   \multicolumn{6}{c|}{} & 80,326 &  \\\cline{1-2}\cline{15-16}

\end{tabular}
}
\end{center}
\caption{Fine-grained sentence-level statistics of the original corpus (a) and the balanced corpus (b) with its five versions.}
\label{tab:sentence-level-stats-fg}
\end{table*}

\clearpage

\begin{table*}[!h]
\begin{center}
\scalebox{0.7}{

\begin{tabular}{|rr|c|c|  ccc  |c|c|c|c|c|rr|}
\multicolumn{2}{c}{} & \multicolumn{2}{c}{\bf(a)} & & & \multicolumn{6}{c}{\bf (b)} & \multicolumn{2}{c}{} \\
\cline{3-4}\cline{8-12}
\multicolumn{2}{c|}{} & \multicolumn{2}{c|}{\bf Original Corpus} & & & & \multicolumn{5}{c|}{\bf Balanced Corpus} & \multicolumn{2}{c}{} \\\cline{1-4}\cline{8-14}

\multicolumn{2}{|c|}{\bf Words}  &\bf  Label &  \multicolumn{1}{c|}{\bf Reinflection Label}   &  & & &  \bf Input & \bf \bf Target\textsubscript{\textit{MM}} &  \bf Target\textsubscript{\textit{FM}} & \bf \bf Target\textsubscript{\textit{MF}} &  \bf \bf Target\textsubscript{\textit{FF}} & \multicolumn{2}{c|}{\bf Words}  \\\cline{1-4}\cline{8-14}

395,658 & 93.5\% & B & & & & & B & B & B & B &  B & 538,733 &  90.3\% \\\hhline{*{4}{=}*{3}{~}*{7}{=}}
1,511 & 0.4\% & 1F & 1M & & & & 1F & 1M & 1F & 1M &  1F & 4,868 &  0.8\% \\\cline{1-4}\cline{8-14}
2,678 & 0.6\% & 1M & 1F & & & & 1M & 1M & 1F & 1M &  1F & 4,868 & 0.8\% \\\cline{1-4}\cline{8-14}
0 & 0\% & 1F! & 1M! & & & & 1F! & 1M! & 1F! & 1M! &  1F! & 55 &  0.01\%\\\cline{1-4}\cline{8-14}
38 & 0.01\% & 1M! & 1F! & & & & 1M! & 1M! & 1F! & 1M! &  1F! & 55 & 0.01\%  \\\hhline{*{4}{=}*{3}{~}*{7}{=}}
6,756 & 1.6\% & 2F & 2M & & & & 2F & 2M & 2M & 2F &  2F & 21,406 & 3.6\% \\\cline{1-4}\cline{8-14}
14,004 & 3.3\% & 2M & 2F & & & & 2M & 2M & 2M & 2F &  2F & 21,406 & 3.6\% \\\cline{1-4}\cline{8-14}
88 & 0.02\% & 2F! & 2M! & & & & 2F! & 2M! & 2M! & 2F! &  2F! & 2,704 & 0.5\% \\\cline{1-4}\cline{8-14}
2,521 & 0.6\% & 2M! & 2F! & & & & 2M! & 2M! & 2M! & 2F! &  2F! & 2,704 & 0.5\% \\\cline{1-4}\cline{8-14}
423,254 &  & \multicolumn{10}{c|}{} & 596,799 &  \\\cline{1-2}\cline{13-14}

\end{tabular}
}
\end{center}
\caption{Fine-grained word-level statistics of the original corpus (a) and the balanced corpus (b) with its five versions.}
\label{tab:word-level-stats-fg}
\end{table*}

\section{Gender Specific English Nouns}
\label{sec:english-gendered-nouns}
Below is the list of  English gender-specific noun lemmas annotated in our dataset. We discard misspelled lemmas in this list. 


\paragraph{Masculine Nouns}
actor, 
boy, boyfriend, bro, brother, businessman, chairman, chap, chauffeur, congressman, cornerman, cowboy, dad, daddy, doorman, emperor, father, footman, foreman, freedman, gentleman, godfather, grandad, granddaddy, 
grandfather, grandpa, grandson, guy, hangman, henchman, highwayman, homeboy, imam, landlord, lawman, lord, lordship, male, man, mate, milkman, nephew, oldman, papa, policeman, pop, praetor, priest, prince, prophet, salesman, samurai, sir, son, stepbrother, uncle, waiter, wizard.

\paragraph{Feminine Nouns}
actress, ballerina, bride, businesswoman, cow, 
daughter, gal, girl, girlfriend, girlie, goddess, godmother, granddaughter, 
grandma, grandmother, housewife, lady, lesbian, ma'am, madam, mama, mom, momma, mommy, mother, mum, nana, nanny, niece, patroness, prima, queen, schoolgirl, sister, sorceress, stepmom, suffragette, supergirl, tsarina, waitress, widow, wife, wingwoman, witch, woman.


\end{document}